\documentclass[11pt,letterpaper]{article}

\usepackage[margin=1in]{geometry}
\usepackage[T1]{fontenc}
\usepackage[utf8]{inputenc}
\usepackage{mathptmx}
\usepackage{courier}
\usepackage{amsmath,amssymb}
\usepackage{graphicx}
\usepackage{booktabs}
\usepackage{microtype}
\usepackage{algorithm}
\usepackage{algpseudocode}
\usepackage[numbers,square]{natbib}
\usepackage[font=small,labelfont=bf]{caption}
\usepackage[hidelinks]{hyperref}

\hypersetup{
  pdftitle={Mask-Pattern Position Augmentation: Training-Free Long-Context Extrapolation for Diffusion LLMs},
  pdfauthor={Gaurav Sarkar and Syed Affan Daimi}
}
\newcommand{\Ltrain}{L_{\mathrm{train}}}
\newcommand{\code}[1]{\texttt{#1}}
\newcommand{\ind}{\mathbf{1}}
\DeclareMathOperator{\clip}{clip}

\title{Mask-Pattern Position Augmentation: Training-Free\\
Long-Context Extrapolation for Diffusion LLMs}
\author{\normalsize Gaurav Sarkar\textsuperscript{1} \qquad Syed Affan Daimi\textsuperscript{1}\\[6pt]
\small \textsuperscript{1}International Institute of Information Technology, Hyderabad}
\date{}

\begin{document}
\maketitle

\begin{abstract}
Diffusion large language models (dLLMs) generate by iteratively unmasking tokens
with bidirectional attention, but rotary position embeddings (RoPE) still confine
them to a fixed pretraining window. Existing training-free extrapolation, notably
LongLLaDA, copies NTK-based RoPE scaling from autoregressive (AR) models
and applies one static, length-only transform at every position and denoising step.
That transplant ignores a signal unique to dLLMs: the evolving mask pattern of the
partially decoded sequence. We propose \textbf{Mask-Pattern Position Augmentation}
(MPPA). MPPA linearly interpolates indices into the pretrained window, then
adds a per-position, per-step offset from cumulative mask density. The offset
restores rotary phase resolution that interpolation collapses, and it shrinks as
tokens are committed, so late steps see near-clean interpolated geometry. On
LLaDA-8B-Instruct at twice its 4096-token window, MPPA reduces perplexity
from 19.51 under standard decoding to 3.12 at 8192 tokens, below the in-window
figure, with 100\% passkey retrieval and unchanged needle-in-a-haystack accuracy.
These preliminary $2\times$ results motivate mask-conditioned position engineering as a
diffusion-native path to long-context generation.
\end{abstract}

\section{Introduction}
\label{sec:introduction}

Masked diffusion language models generate text by denoising: starting from a fully masked answer,
a bidirectional transformer repeatedly predicts masked tokens and re-masks the least confident ones
until the sequence is clean \citep{austin2021,lou2024,sahoo2024,nie2025}. This decouples the number of decoding steps from sequence
length and gives every position access to both left and right context, properties that make dLLMs an
attractive non-autoregressive alternative. Like AR transformers, however, open dLLMs encode order
with rotary position embeddings \citep{su2021} trained on a bounded window $\Ltrain$, and their quality degrades
once inference exceeds that window.

Liu et al.\ \citep{liu2025} give the first systematic study of long-context dLLMs. They report two findings: dLLMs
keep remarkably stable perplexity under direct context extrapolation (where AR models diverge), and
they exhibit a \emph{local perception} phenomenon, retrieving reliably only from context near the generation
window. Their method, LongLLaDA, imports NTK-aware RoPE scaling \citep{peng2023} into dLLM inference
and shows that AR extrapolation scaling laws transfer. Crucially, this transplant treats the dLLM
inference loop as if it were AR: RoPE is rescaled once, by a factor that depends only on the target
length, and applied identically at every denoising step and every position.

We argue this leaves the most distinctive property of dLLMs unused. At every denoising step the
model is handed an explicit \textbf{mask pattern}: which positions are still masked and which have been
committed. This pattern is a first-class input to a dLLM (AR decoders never see it) and it changes
as denoising proceeds: dense with masks early, nearly clean late. It is, in effect, a free, monotone,
step-varying signal aligned with the generation process. MPPA uses it to shape positions.

Concretely, MPPA (i) linearly interpolates positions into $[0,\Ltrain)$ to keep rotary angles in-distribution,
then (ii) adds a small offset proportional to the cumulative mask density at each position.
Because the offset is largest exactly where masks concentrate (the tokens being generated) and shrinks
as they are filled in, it acts as a \textbf{denoising-synchronized dither}: it re-separates rotary phases that
interpolation would otherwise collapse (the interpolation degeneracy problem \citep{chen2023}) early in decoding,
and it vanishes for the final refinement steps. MPPA is training-free, adds one prefix-sum per forward
pass, and is a drop-in replacement for the position indices.

\paragraph{Contributions.}
\begin{itemize}
  \item We identify the mask pattern as a diffusion-native, denoising-varying signal for position engineering,
  unavailable to AR length-extrapolation methods (\S\ref{sec:method}).
  \item We propose MPPA: position interpolation plus mask-density position dithering, with a RoPE-degeneracy
  rationale for why a mask-conditioned, self-annealing offset restores the positional
  resolution that interpolation destroys (\S\ref{sec:method}).
  \item In a preliminary $2\times$ study on LLaDA-8B-Instruct, MPPA reduces 8192-token perplexity by $6.3\times$
  over standard decoding, to below its in-window value, while retaining perfect passkey retrieval
  and unchanged needle retrieval (\S\ref{sec:experiments}).
\end{itemize}
We position MPPA as an efficiency method for training-free long-context generation,
with a mechanism grounded in the foundations of RoPE geometry.

\section{Background and positioning}
\label{sec:background}

\paragraph{RoPE and interpolation.}
RoPE rotates each query/key feature pair $j\in\{0,\ldots,d/2-1\}$ at
position $p$ by an angle $\theta_{p,j}=p\,b^{-2j/d}$, where $b$ is the base (\code{rope\_theta}) \citep{su2021}; attention depends
only on relative phases $(p-p')\,b^{-2j/d}$. Two training-free ways to fit a longer sequence into the
trained angular range are position interpolation (PI), which rescales indices $p\mapsto p\,\Ltrain/L$ \citep{chen2023}, and
NTK/YaRN scaling, which enlarges the base $b$ so low frequencies stretch while high frequencies
are nearly preserved \citep{peng2023}. Self-Extend instead groups distant positions with integer floor division
at inference \citep{jin2024}. PI is simple but suffers \emph{interpolation degeneracy}: for large $L$ the per-index phase
increment $\Ltrain/L\cdot b^{-2j/d}$ shrinks, so adjacent positions map to nearly identical high-frequency
phases and the model loses local order resolution.

\paragraph{Long-context dLLMs.}
LongLLaDA \citep{liu2025} is the closest work: it is training-free, targets LLaDA \citep{nie2025},
and evaluates with perplexity, passkey retrieval, needle-in-a-haystack (NIAH) \citep{li2024}, and RULER \citep{hsieh2024}.
Its lever is NTK scaling. All prior levers (PI, NTK/YaRN, Self-Extend) are functions of position
indices only. MPPA's point of departure is to condition the position transform on the \textbf{mask state} $x^{(t)}$,
which is specific to iterative masked decoding and changes with the denoising step $t$.

\section{Method: MPPA}
\label{sec:method}

\paragraph{Setup.}
We use LLaDA-style semi-autoregressive block decoding: the answer of length $G$ is split
into blocks; within a block the model runs $S$ denoising steps, each predicting all masked tokens,
scoring confidence, and committing a scheduled number of the most confident ones (low-confidence
remasking) \citep{nie2025}. Let $x^{(t)}\in\mathcal{V}^{L}$ be the sequence at step $t$ over a prompt of length $P$ and total length
$L=P+G$, and let $M$ denote the mask id.

\paragraph{Mask-density signal.}
For each position $i$ define the cumulative fraction of masked tokens up to $i$,
\begin{equation}
  m_i^{(t)}=\frac{1}{i+1}\sum_{k=0}^{i}\ind\bigl[x_k^{(t)}=M\bigr]\in[0,1].
  \label{eq:mask-density}
\end{equation}
Over a clean prompt $m_i^{(t)}\approx0$; over the still-masked generation region it is large early in decoding
and decreases monotonically as tokens are committed.

\begin{algorithm}[tbp]
  \caption{MPPA block decoding. The mask-density signal $m$ and augmented positions $\tilde p$ (lines 4 to 6) are the only additions to standard LLaDA block decoding.}
  \label{alg:mppa}
  \small
  \begin{algorithmic}[1]
    \Require prompt $x_{0:P}$; block length $B$; steps $S$; strength $\alpha$; window $\Ltrain$; mask id $M$
    \State initialize $x\gets[x_{0:P},M^G]$; $L\gets P+G$; $c_L\gets\min(1,\Ltrain/L)$
    \For{denoising step $t=1,\ldots,S$}
      \State $\mathrm{mask}_i\gets\ind[x_i=M]$ for all $i$ \Comment{current mask pattern}
      \State $m_i\gets\frac{1}{i+1}\sum_{k\le i}\mathrm{mask}_k$
      \Statex \hfill $\triangleright$ cumulative mask density (prefix sum, $O(L)$)
      \State $\tilde p_i\gets\clip\!\left(\left\lfloor c_Li+\alpha m_i\frac{1}{10}\min(\Ltrain,L)\right\rfloor,\,0,\,\max(\Ltrain,L)-1\right)$
      \Statex \hfill $\triangleright$ augmented positions
      \State $\ell\gets\operatorname{model}(x,\tilde p)$
      \Statex \hfill $\triangleright$ frozen weights and \code{rope\_theta}; feed $\tilde p$ as \code{position\_ids}
      \State commit the scheduled top-confidence masked tokens of $\ell$
      \Statex \hfill $\triangleright$ low-confidence remasking
    \EndFor
    \State \Return $x$
  \end{algorithmic}
\end{algorithm}

\paragraph{Augmented positions.}
With interpolation factor $c_L=\min(1,\Ltrain/L)$ and strength $\alpha$, MPPA
replaces the standard indices by
\begin{equation}
  \begin{aligned}
    \tilde p_i^{(t)}
      &=\left\lfloor
        \underbrace{c_Li}_{\text{PI base}}
        +\underbrace{\alpha m_i^{(t)}\frac{1}{10}\min(\Ltrain,L)}_{\text{mask-density dither}}
        \right\rfloor,\\[2pt]
    \tilde p_i^{(t)}
      &\gets\clip\!\left(\tilde p_i^{(t)},\,0,\,\max(\Ltrain,L)-1\right),
  \end{aligned}
  \label{eq:augmented-positions}
\end{equation}
and feeds $\tilde p^{(t)}$ as \code{position\_ids} to the frozen model. The base term keeps angles in-distribution;
the dither term shifts positions forward in proportion to how mask-dominated their prefix is. It is
therefore concentrated on the generation region and recomputed every step, so it \emph{co-varies with
denoising}. Algorithm~\ref{alg:mppa} shows one block.

\paragraph{Why it works.}
Under PI, positions $i$ and $i+1$ differ in phase by $c_L\,b^{-2j/d}$, which is small for
large $L$: neighboring generation tokens become hard to order (degeneracy). MPPA adds a data-dependent
term $\delta_i^{(t)}=\alpha m_i^{(t)}\frac{1}{10}\min(\Ltrain,L)$ whose \emph{differences} across positions re-separate these
phases exactly where masks live. Two mechanisms follow. (i) \emph{Degeneracy breaking}: collapsed
high-frequency phases in the generation region regain distinguishable rotations. (ii) \emph{Denoising
annealing}: as decoding commits tokens, $m^{(t)}$ falls, so $\delta^{(t)}\to0$ and the final refinement steps see
near-clean interpolated positions: strong dithering for early global structure, gentle geometry for late
local polishing. Both are only definable because a dLLM ingests an explicit, evolving mask pattern.
The overhead is one prefix sum per step; weights and \code{rope\_theta} are untouched.

\section{Experiments}
\label{sec:experiments}

\paragraph{Setup.}
We evaluate on GSAI-ML/LLaDA-8B-Instruct with $\Ltrain=4096$ and its native
\code{rope\_theta}$=5\times10^5$, comparing at the trained length (4096) and at $2\times$ extrapolation (8192). Decoding
is identical in both settings (\code{steps}$=32$, \code{block}$=32$, temperature 0, low-confidence remasking);
MPPA uses $\alpha=0.1$. Three tasks follow LongLLaDA: Monte-Carlo perplexity (PPL) on a government-report
passage (lower is better); synthetic passkey retrieval; and needle-in-a-haystack (NIAH) on
NeedleBench text with keyword/edit-distance scoring \citep{li2024}. Passkey/NIAH are averaged over insertion
depths $\{0,25,50,75,100\}\%$. Our comparator is the STANDARD decoder run unchanged (identity
position indices and native \code{rope\_theta}), i.e. direct context extrapolation \citep{liu2025}; MPPA differs from it
only by the position map of Algorithm~\ref{alg:mppa}, isolating the effect of mask-conditioned positions.

\begin{table}[tbp]
  \centering
  \caption{LLaDA-8B-Instruct at the trained window (4096) and $2\times$ extrapolation (8192). PPL is Monte-Carlo perplexity (lower is better); passkey and NIAH are accuracy averaged over five insertion depths (higher is better). The decisive regime is 8192: at 4096 the model operates within its trained window.}
  \label{tab:results}
  \small
  \setlength{\tabcolsep}{5pt}
  \begin{tabular}{lrrrrrr}
    \toprule
      & \multicolumn{2}{c}{PPL $\downarrow$}
      & \multicolumn{2}{c}{Passkey (\%) $\uparrow$}
      & \multicolumn{2}{c}{NIAH (\%) $\uparrow$} \\
    \cmidrule(lr){2-3}\cmidrule(lr){4-5}\cmidrule(lr){6-7}
    Method & 4096 & 8192 & 4096 & 8192 & 4096 & 8192 \\
    \midrule
    Standard decoding (no rescaling) & 2.58 & 19.51 & 100 & 100 & 100 & 60 \\
    \textbf{MPPA (ours)} & 5.62 & \textbf{3.12} & 100 & 100 & 100 & 60 \\
    \bottomrule
  \end{tabular}
\end{table}

\begin{figure}[tbp]
  \centering
  \includegraphics[width=0.85\linewidth]{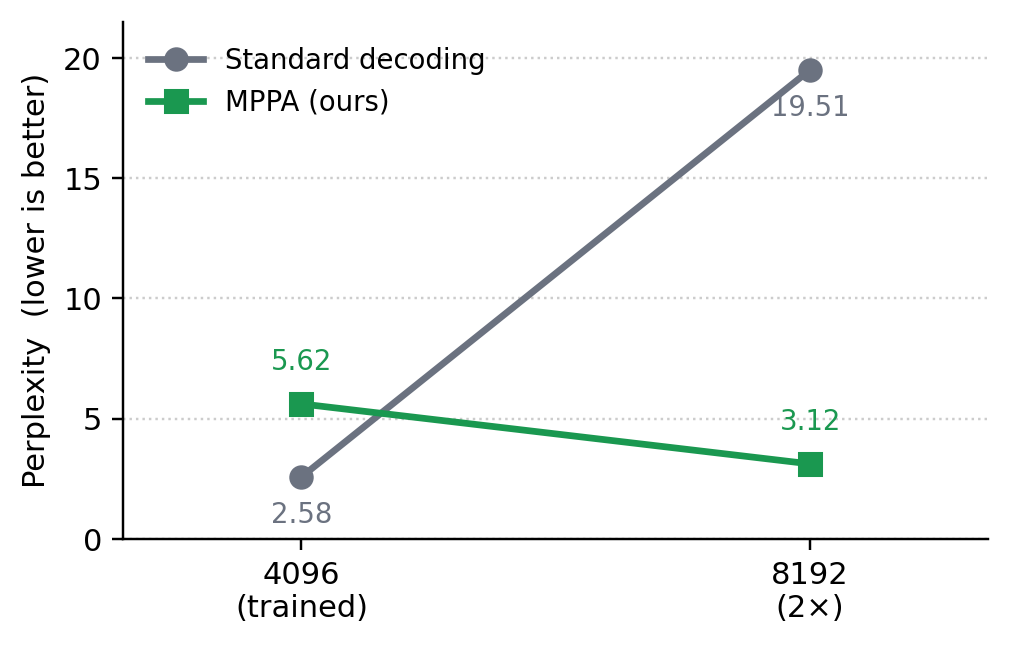}
  \caption{Perplexity vs. context length (plotted from the benchmark log). Standard decoding collapses under $2\times$ direct extrapolation ($2.58\to19.51$); MPPA instead decreases to 3.12, below its own in-window value.}
  \label{fig:perplexity}
\end{figure}

\begin{figure}[tbp]
  \centering
  \includegraphics[width=0.75\linewidth]{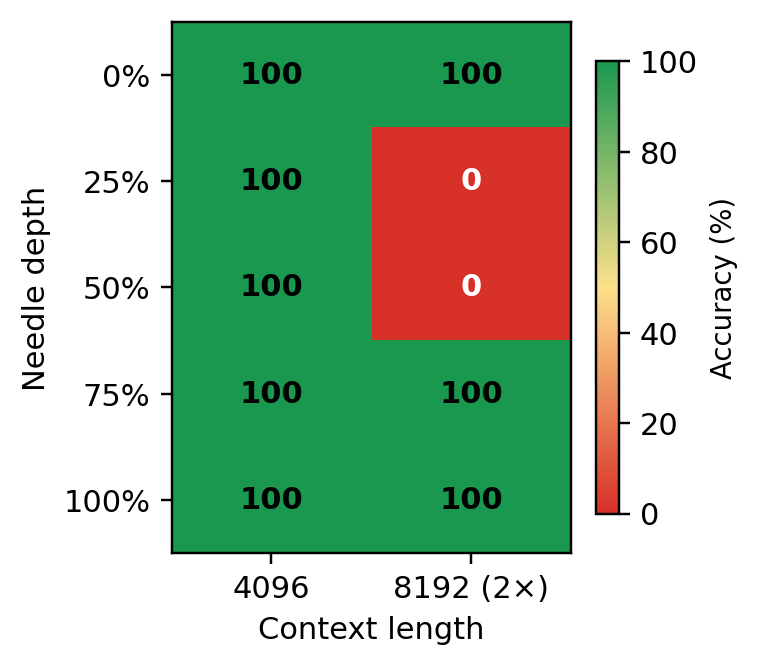}
  \caption{MPPA needle-in-a-haystack accuracy by insertion depth and context length. Retrieval is perfect within the trained window and at the shallow/deep ends at $2\times$; only the 25 to 50\% mid-depths fail at 8192, matching LLaDA's local-perception behavior.}
  \label{fig:niah}
\end{figure}

\paragraph{Perplexity.}
Table~\ref{tab:results} and Figure~\ref{fig:perplexity} isolate the $2\times$ regime the method targets (4096 is in-window, where
the unmodified decoder is naturally strongest). Under direct extrapolation the standard decoder's
perplexity rises from 2.58 to 19.51 as the sequence leaves the trained window. MPPA instead lowers
it to 3.12 at 8192, a $6.3\times$ reduction, and below its own 4096 value of 5.62. That perplexity \emph{improves}
rather than degrades from 4096 to 8192 is consistent with the dither counteracting interpolation
degeneracy exactly when interpolation becomes active ($L>\Ltrain$); within the trained window the
always-on offset perturbs already in-distribution positions, so the in-window figure is a mild cost
rather than a gain.

\paragraph{Retrieval.}
MPPA recovers the hidden passkey in 100\% of trials at every depth for both lengths,
and its needle accuracy is unchanged from standard decoding: perfect at 4096 and 60\% at 8192
(Figure~\ref{fig:niah}). The residual $2\times$ loss is confined to the 25 to 50\% mid-depths, where standard decoding
also fails at 8192; this reflects the model's local-perception bias rather than a positional defect MPPA
can remove alone. The takeaway is that MPPA's large perplexity gain comes at no retrieval cost.

\section{Discussion, limitations, and conclusion}
\label{sec:discussion}

\paragraph{Limitations.}
This is a preliminary study at a single $2\times$ extrapolation on one model; 16k to 32k
lengths and the RULER/LongBench suites used by LongLLaDA are the immediate next step. The
mid-depth NIAH ceiling (25 to 50\% depths fail at 8192) reflects LLaDA's local-perception behavior
rather than a positional artifact MPPA can fix alone. Because the offset is always on, within the
trained window it perturbs in-distribution positions and its in-window PPL (5.62) trails standard
decoding (2.58); a gated variant that disables the dither when $L\le\Ltrain$ is an obvious fix. Finally, $\alpha$
and the $\frac{1}{10}$ scale are untuned defaults, retrieval uses one trial per depth, and PPL uses few Monte-Carlo
samples; sensitivity sweeps and error bars are pending.

\paragraph{Conclusion.}
MPPA shows that the mask pattern, an input dLLMs have and AR models lack, is a
useful, essentially free control signal for training-free length extrapolation. Interpolating positions
and then dithering them by a self-annealing, mask-density offset cuts $2\times$ perplexity by $6.3\times$ over
standard decoding on LLaDA-8B while preserving retrieval, and it does so with a single prefix sum
and no weight or \code{rope\_theta} changes. We hope this motivates position (and attention) transforms
that are conditioned on the denoising state rather than transplanted unchanged from autoregressive
models.

\bibliographystyle{plainnat}
\bibliography{references}

\begin{thebibliography}{11}
\providecommand{\natexlab}[1]{#1}
\providecommand{\url}[1]{\texttt{#1}}
\expandafter\ifx\csname urlstyle\endcsname\relax
  \providecommand{\doi}[1]{doi: #1}\else
  \providecommand{\doi}{doi: \begingroup \urlstyle{rm}\Url}\fi

\bibitem[Austin et~al.(2021)Austin, Johnson, Ho, Tarlow, and van~den
  Berg]{austin2021}
J.~Austin, D.~D. Johnson, J.~Ho, D.~Tarlow, and R.~van~den Berg.
\newblock Structured denoising diffusion models in discrete state-spaces.
\newblock In \emph{NeurIPS}, 2021.

\bibitem[Chen et~al.(2023)Chen, Wong, Chen, and Tian]{chen2023}
S.~Chen, S.~Wong, L.~Chen, and Y.~Tian.
\newblock Extending context window of large language models via positional
  interpolation.
\newblock \emph{arXiv:2306.15595}, 2023.

\bibitem[Hsieh et~al.(2024)Hsieh, Sun, Kriman, et~al.]{hsieh2024}
C.-P. Hsieh, S.~Sun, S.~Kriman, et~al.
\newblock {RULER}: What's the real context size of your long-context language
  models?
\newblock \emph{arXiv:2404.06654}, 2024.

\bibitem[Jin et~al.(2024)Jin, Han, Yang, et~al.]{jin2024}
H.~Jin, X.~Han, J.~Yang, et~al.
\newblock {LLM} maybe {LongLM}: {Self-Extend LLM} context window without
  tuning.
\newblock In \emph{ICML}, 2024.

\bibitem[Li et~al.(2024)Li, Zhang, Liu, et~al.]{li2024}
M.~Li, S.~Zhang, Y.~Liu, et~al.
\newblock {NeedleBench}: Can {LLMs} do retrieval and reasoning in {1M} context
  window?
\newblock \emph{arXiv:2407.11963}, 2024.

\bibitem[Liu et~al.(2025)Liu, Song, Liu, Huang, Guo, He, and Qiu]{liu2025}
X.~Liu, Y.~Song, Z.~Liu, Z.~Huang, Q.~Guo, Z.~He, and X.~Qiu.
\newblock {LongLLaDA}: Unlocking long context capabilities in diffusion {LLMs}.
\newblock \emph{arXiv:2506.14429}, 2025.

\bibitem[Lou et~al.(2024)Lou, Meng, and Ermon]{lou2024}
A.~Lou, C.~Meng, and S.~Ermon.
\newblock Discrete diffusion modeling by estimating the ratios of the data
  distribution.
\newblock In \emph{ICML}, 2024.

\bibitem[Nie et~al.(2025)Nie, Zhu, You, et~al.]{nie2025}
S.~Nie, F.~Zhu, Z.~You, et~al.
\newblock Large language diffusion models.
\newblock \emph{arXiv:2502.09992}, 2025.

\bibitem[Peng et~al.(2023)Peng, Quesnelle, Fan, and Shippole]{peng2023}
B.~Peng, J.~Quesnelle, H.~Fan, and E.~Shippole.
\newblock {YaRN}: Efficient context window extension of large language models.
\newblock \emph{arXiv:2309.00071}, 2023.

\bibitem[Sahoo et~al.(2024)Sahoo, Arriola, Schiff, et~al.]{sahoo2024}
S.~S. Sahoo, M.~Arriola, Y.~Schiff, et~al.
\newblock Simple and effective masked diffusion language models.
\newblock In \emph{NeurIPS}, 2024.

\bibitem[Su et~al.(2021)Su, Lu, Pan, Wen, and Liu]{su2021}
J.~Su, Y.~Lu, S.~Pan, B.~Wen, and Y.~Liu.
\newblock {RoFormer}: Enhanced transformer with rotary position embedding.
\newblock \emph{arXiv:2104.09864}, 2021.

\end{thebibliography}

\clearpage
\appendix
\section{Reproducibility details}
\label{app:reproducibility}

\paragraph{Model and decoding.}
GSAI-ML/LLaDA-8B-Instruct in bfloat16, SDPA attention, native
\code{rope\_theta}$=5\times10^5$, $\Ltrain=4096$. Shared decoding: \code{steps}$=32$, \code{block\_length}$=32$, generation
length 32 (passkey) / 64 (NIAH), temperature 0, low-confidence remasking. Seed 2025.
MPPA: $\alpha=0.1$, offset scale $\frac{1}{10}\min(\Ltrain,L)$, positions clipped to $[0,\max(\Ltrain,L)-1]$, all
other kwargs shared.

\paragraph{Tasks.}
PPL: Monte-Carlo masked cross-entropy on a $\sim$16k-token government-report passage,
chunk size 64, 4 MC samples, exponentiated. Passkey: a 5-digit key hidden in filler at depths
$\{0,25,50,75,100\}\%$; success iff the key appears in the decoded answer. NIAH: NeedleBench
\code{PaulGrahamEssays} haystack with English needles; keyword hit scores 100, otherwise $0.2\times$ the
normalized edit-distance score; here reported as accuracy averaged over depths. All numbers are
from a single benchmark run and should be read as preliminary.

\paragraph{Methods.}
STANDARD decoding uses identity position indices with the native \code{rope\_theta} (direct
extrapolation). MPPA replaces only the position indices with $\tilde p^{(t)}$ from Algorithm~\ref{alg:mppa} ($\alpha=0.1$);
weights, \code{rope\_theta}, and the decoding schedule are otherwise identical. Both are training-free and
require no weight updates.

\end{document}